%% file: acl_latex.tex
\pdfoutput=1

\documentclass[11pt]{article}

\usepackage[preprint]{acl}

\usepackage{times}
\usepackage{latexsym}

\usepackage[T1]{fontenc}

\usepackage[utf8]{inputenc}

\usepackage{microtype}

\usepackage{inconsolata}

\usepackage{graphicx}

\usepackage{xspace}
\usepackage{multirow} 
\usepackage{amsmath}
\usepackage{graphicx}  
\usepackage{caption}  
\usepackage{tabularray}
\usepackage{siunitx} 
\usepackage{amssymb}
\newcommand{\benchmark}{\textsc{Code-Vision}\xspace}
\usepackage{booktabs}
\usepackage[draft]{minted}
\usepackage{longtable}
\title{\benchmark: Evaluating Multimodal LLMs Logic Understanding and Code Generation Capabilities}

\author{Hanbin Wang$^{1}$\thanks{ \ \ indicates equal contribution.}, Xiaoxuan Zhou$^{2*}$, Zhipeng Xu$^{2}$, Keyuan Cheng$^{1}$, Yuxin Zuo$^{3}$, \\ \textbf{Kai Tian$^{2}$, Jingwei Song$^{4}$, Junting Lu$^{1}$, Wenhui Hu$^{1}$ and Xueyang Liu$^{1}$} \\ 
$^1$Peking University $^2$Northeastern University \\
$^3$Institute of Computing Technolgy, Chinese Academy of Sciences 
$^4$University of Hong Kong \\
}

\begin{document}
\maketitle
\begin{abstract}
This paper introduces \benchmark, a benchmark designed to evaluate the logical understanding and code generation capabilities of Multimodal Large Language Models (MLLMs). It challenges MLLMs to generate a correct program that fulfills specific functionality requirements based on a given flowchart, which visually represents the desired algorithm or process. \benchmark comprises three subsets—HumanEval-V, Algorithm, and MATH, which evaluate MLLMs' coding abilities across basic programming, algorithmic, and mathematical problem-solving domains. Our experiments evaluate 12 MLLMs on \benchmark. Experimental results demonstrate that there is a large performance difference between proprietary and open-source models. On Hard problems, GPT-4o can achieve 79.3\% pass@1, but the best open-source model only achieves 15\%. Further experiments reveal that \benchmark can pose unique challenges compared to other multimodal reasoning benchmarks MMCode and MathVista. We also explore the reason for the poor performance of the open-source models. All data and codes are available at \url{https://github.com/wanghanbinpanda/CodeVision}. 
\end{abstract}

\input{sec/1_intro}

\input{sec/2_related_work}

\input{sec/3_dataset}

\input{sec/4_experimental_methodology}

\input{sec/5_evaluation_results}

\input{sec/6_conclusion}
\section*{Limitations}
\input{sec/7_Limitations}

\section*{Ethics Statement}
\input{sec/8_Ethics}

\bibliography{custom}

\clearpage
\newpage
\appendix

\input{sec/X_suppl}

\end{document}

%% file: sec/1_intro.tex
\section{Introduction}
\input{figures/motivation}
Recent years have witnessed remarkable progress in Multimodal Large Language Models (MLLMs), which can process and generate information across different modalities such as text, images, and code~\cite{gpt4o,Claude3,Claude3.5_Sonnet,Llama3.2,geminiteam2024gemini15unlockingmultimodal,yao2024minicpm}. These models such as GPT-4o~\cite{gpt4o}, Claude-3~\cite{Claude3}, Gemini~\cite{geminiteam2024gemini15unlockingmultimodal}, and Llama-3.2-Vision~\cite{Llama3.2} have demonstrated impressive capabilities in various tasks, including visual question answering~\cite{singh2019towards,goyal2017making,marino2019ok,li2021adversarial}, mathematical reasoning~\cite{lu2024mathvista,wang2024measuring}, and code generation~\cite{li2024mmcode,shi2024chartmimic}. The ability to understand and reason about visual information while generating accurate code is particularly crucial for advancing artificial intelligence systems.  

However, existing benchmarks for evaluating MLLMs' logic understanding and code generation capabilities have significant limitations. While benchmarks like MMCode~\cite{li2024mmcode} have contributed valuable insights, they often treat visual information as supplementary rather than essential. For instance, as shown in Figure \ref{fig:motivation}, in MMCode, many programming problems can be solved based solely on text descriptions without requiring visual understanding. This limitation makes it difficult to assess whether MLLMs truly utilize and comprehend visual information in their coding process. Meanwhile, when humans solve problems in mathematics and programming, they often use flowcharts to visualize the logic and structure of the problem. If MLLMs can translate flowcharts into code accurately, it will greatly increase the efficiency of programming and problem solving. However, MLLMs has not been evaluated in this scenario.

To address these challenges, we introduce \benchmark, a novel benchmark designed to evaluate the logical understanding and code generation capabilities of MLLMs. \benchmark differs from existing benchmarks in several key aspects: 1) \textbf{Visual-Centric Design}: Our benchmark uses flowcharts as the primary input, making visual information essential rather than supplementary. Without understanding the flowchart, it is nearly impossible for models to generate correct code solutions. 2) \textbf{Comprehensive Evaluation}: \benchmark comprises three distinct subsets—HumanEval-V, Algorithm, and MATH, which evaluate MLLMs' reasoning abilities across basic programming, algorithmic, and mathematical problem-solving domains. 3) \textbf{Rigorous Testing}: Each problem in \benchmark includes comprehensive test cases covering typical scenarios, edge cases, and large-number inputs, ensuring a thorough evaluation of the generated code's correctness and robustness.

Through extensive experiments with 12 state-of-the-art MLLMs, we demonstrate that \benchmark effectively reveals significant performance differences between proprietary and open-source models. Our results show that while leading proprietary models like GPT-4o can achieve up to 79.3\% pass@1 on hard problems, the best open-source models struggle to surpass 15\%. All open-source models consistently fail to solve problems in the Algorithm Hard category, achieving a 0\% pass rate. These findings highlight the continuing challenges in developing open-source MLLMs that can match the reasoning capabilities of proprietary models.

Our further analyses also reveal interesting insights. When comparing with MathVista~\cite{lu2024mathvista}, we find that proprietary models maintain similar performance across both benchmarks, while open-source models show a significant performance drop (around -30\%) on \benchmark, highlighting the unique challenges our benchmark poses for testing algorithmic reasoning capabilities. Our error analysis demonstrates a clear distinction in failure patterns: proprietary models primarily fail due to logical errors (AssertionError) while maintaining syntactic correctness, whereas open-source models frequently struggle with basic code structure and syntax issues, suggesting fundamental gaps in their code generation abilities. These findings provide valuable insights into the current limitations of MLLMs in visual-based code generation tasks.

%% file: figures/motivation.tex
\begin{figure}[t] \centering
    \includegraphics[width=0.48\textwidth]{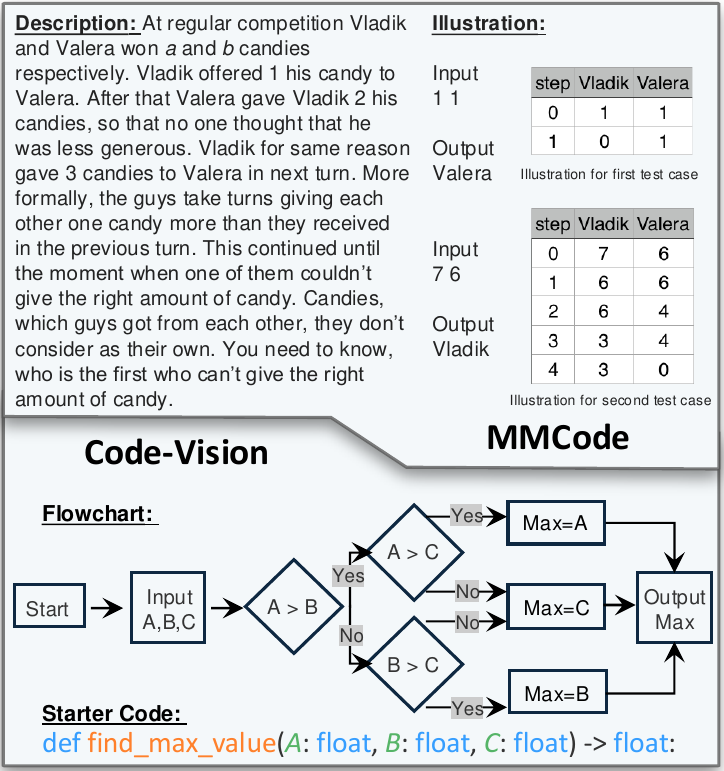}
    \caption{Comparison of data examples from \benchmark and MMCode. In MMCode, images serve a supplementary role, while in \benchmark, images play a primary role.} \label{fig:motivation}
\end{figure}

%% file: sec/2_related_work.tex
\section{Related Work}
\subsection{Multimodal LLMs} 
In recent years, Multimodal Large Language Models (MLLMs) have gained significant attention due to their ability to process and generate information across various modalities, such as text, images, and audio~\cite{li2024llava,liu2024visual,zhu2023minigpt,li2024llavaonevisioneasyvisualtask,dai2023instructblipgeneralpurposevisionlanguagemodels,ye2024mplugowlmodularizationempowerslarge}. Some proprietary models such as GPT-4o~\cite{gpt4o}, Claude-3~\cite{Claude3}, and Gemini~\cite{geminiteam2024gemini15unlockingmultimodal} show superior performance, especially on visually complex reasoning tasks such as MathVista~\cite{lu2024mathvista}. Open-source MLLMs have made strides as well, notable models include Llama-3.2-Vision~\cite{Llama3.2}, Phi-3-Vision~\cite{abdin2024phi}, MiniCPM V2.6~\cite{yao2024minicpm}, Qwen-VL~\cite{bai2023qwenvlversatilevisionlanguagemodel}, Deepseek-VL~\cite{lu2024deepseek}. The reasoning ability of MLLMs has recently become a highly regarded research focus because they can extend beyond language reasoning to encompass multiple modalities (such as images), achieving more comprehensive and complex reasoning. Previous research has evaluated the reasoning ability of MLLMs through tasks like Visual Question Answering (VQA)~\cite{mobasher2022parsvqa,gurari2018vizwiz} and mathematical reasoning~\cite{lu2024mathvista}. In this paper, we evaluate the logical understanding and reasoning capabilities of MLLMs through code generation.
\input{figures/construction}
·\subsection{Multimodal Reasoning Benchmarks}
The reasoning capabilities of Multimodal LLMs are important for the development of Artificial General Intelligence (AGI)~\cite{morris2023levels}. To advance the reasoning capabilities of MLLMs, the research community has constructed several multi-modal reasoning benchmarks~\cite{li2024survey}. MathVista~\cite{lu2024mathvista} and Math-Vision
~\cite{wang2024measuring} to evaluate the mathematical reasoning abilities of MLLMs within visual contexts. Additionally, inspired by the effectiveness of code in evaluating the reasoning capabilities of LLMs~\cite{chen2021evaluatinglargelanguagemodels,hendrycksapps2021,jain2024livecodebench,austin2021program}, many researchers have begun using code generation to evaluate the reasoning abilities of MLLMs by introducing visual contexts.
Design2Code~\cite{si2024design2code} evaluates the code generation abilities of MLLMs through HTML web page generation, which involves converting User Interfaces (UI) into clean and responsive HTML/CSS/JS code. ChartMimic~\cite{shi2024chartmimic} requires MLLMs to generate corresponding chart rendering code by using information-dense visual charts and text instructions as input. MMCode~\cite{li2024mmcode} is relevant to our work, and it evaluates algorithmic problem-solving skills in visually rich contexts. However, in MMCode, visual information is not dominant, and images only serve as supplementary elements. Many programming competition problems can still be solved based solely on the text description without the images. Inspired by this, we propose \benchmark, where visual information is predominant, meaning that without the images, it is nearly impossible for the model to solve the problem. This allows for a better evaluation of the MLLMs' understanding and reasoning abilities.

%% file: figures/construction.tex
\begin{figure*}[t] \centering
    \includegraphics[width=1\textwidth]{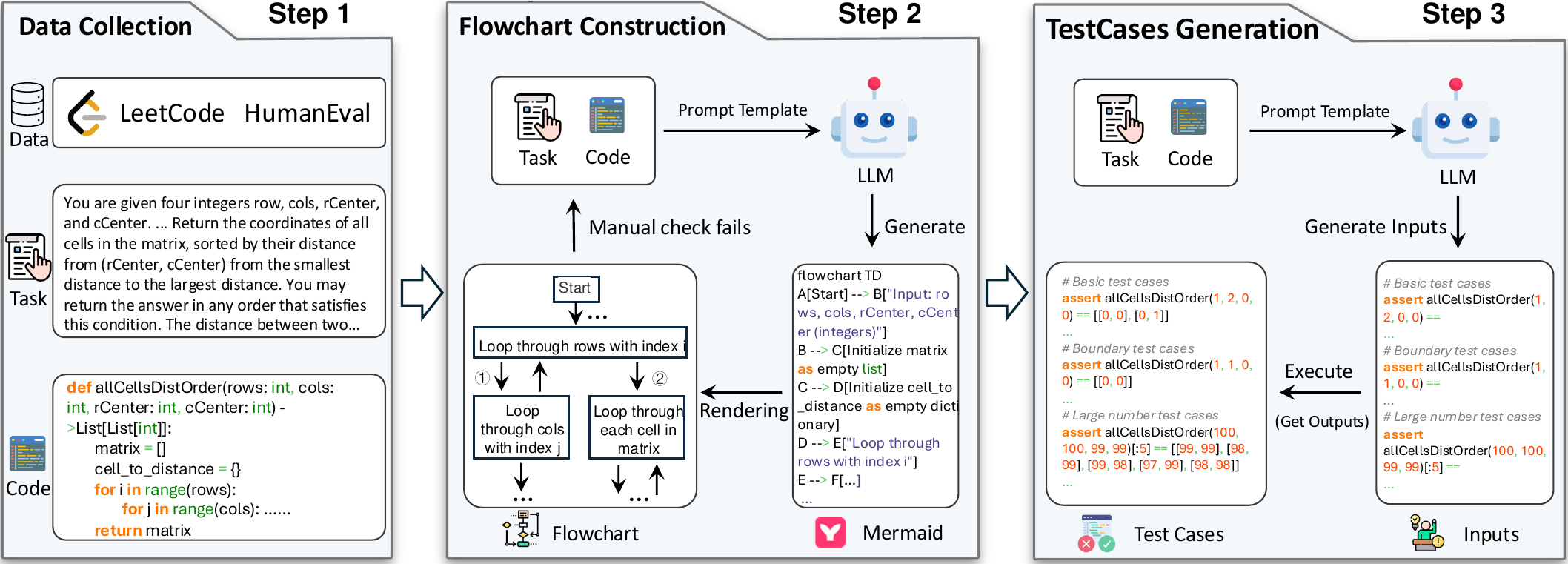}
    \caption{\benchmark Construction Method. The Method includes Data Collection, Flowchart Construction, and Test Cases Generation.} \label{fig:construction}
\end{figure*}

%% file: sec/3_dataset.tex
\section{\benchmark}
In this section, we introduce the benchmark \benchmark, which evaluates the logical understanding and code generation abilities of Multimodal Large Language Models (MLLMs). We first describe how \benchmark tests MLLMs (Sec.~\ref{dataset:definition}). Then we detail the process of constructing the \benchmark benchmark (Sec.~\ref{dataset:collection}) and show the statistics of \benchmark(Sec.~\ref{dataset:sta}).

\subsection{Task Definition}\label{dataset:definition}

In \benchmark, MLLMs require generating a correct program \( P \) that fulfills specific functionality requirements based on a given flowchart \( F \), which visually represents the desired algorithm or process. 

Generating code involves translating visual or descriptive elements from the image into logical and structured code that meets the required specifications. After generating the code \( P \), we evaluate its correctness by using a set of test cases \( X = \{(x_1, y_1), \ldots, (x_n, y_n)\} \). Specifically, we provide input \( x_i \) to the generated code \( P \) and obtain the execution result \( P(x_i) \). If there exists any test case \( (x_i, y_i) \in X \) that satisfies \( P(x_j) \neq y_j \), it indicates that the generated code \( P \) does not fulfill the required functionality. If the MLLM generates a program \( P \) that correctly implements the specifications derived from the flowchart \( F \) for all test cases \( \forall (x_i, y_i) \in X, P(x_i) = y_i \), it demonstrates the MLLM's capability to accurately interpret and convert visual algorithmic instructions into executable code, ensuring reliability and correctness in its outputs. This successful translation confirms the model's understanding of both the structure and logic inherent in the flowchart, marking an important step towards advanced reasoning abilities.

\subsection{Data Construction}\label{dataset:collection}
In this subsection, we outline the methodology behind the construction of the \benchmark dataset. As shown in Figure ~\ref{fig:construction}, it includes Data Collection, Flowchart Construction, and Test Cases Generation.

\textbf{Data Collection.} \benchmark consists of three subsets: HumanEval-V, Algorithm, and MATH, which together provide a comprehensive evaluation of MLLMs' reasoning abilities across three aspect: basic programming, algorithm, and math. To ensure the quality and diversity of \benchmark, we collect data from the LeetCode website\footnote{\url{https://leetcode.com/}} as well as some open source high-quality data such as HumanEval~\cite{chen2021evaluatinglargelanguagemodels}.

HumanEval-V is collected from HumanEval, which is a code generation benchmark specifically designed to assess the capability of LLMs in generating correct and efficient Python code based on natural language prompts. We store all the data from HumanEval in \benchmark. Algorithm and MATH subsets are constructed by carefully curating a variety of problems from the LeetCode website that target specific reasoning skills necessary for proficient programming and mathematical problem-solving. We categorize the problems into three levels of difficulty: easy, medium, and hard, to further refine the assessment of the model's capabilities. In the Algorithm subset, problems at each level cover algorithms such as data structure operations and dynamic programming. In the MATH subset, the problems encompass mathematical concepts such as arithmetic operations, algebraic knowledge, and symbolic manipulation.

\input{tables/stats}

\textbf{Flowchart Construction. } The most important part of \benchmark is to build a clear flow chart that accurately depicts the logical structure of the problem and the steps to solve it. Therefore, when building the flowchart, we use a two-step method of creating a mermaid and then rendering it into a visual flowchart to ensure that the flowchart accurately reflects the logic of the problem.

We first use GPT-4o to generate the mermaid from the problem descriptions and correct code solutions. The prompt we used is in Figure ~\ref{fig:prompt}. Mermaid is a text-based flowchart representation that allows us to describe diagrams and visual data flows in a straightforward manner. After generating the mermaid, we utilize an automated rendering tool\footnote{\url{https://www.mermaidchart.com/}} to render it into a high-quality visualization flowchart. During the process of rendering the flowchart, we manually review the automatically generated flowchart to ensure its compliance and logicality. For those flowcharts whose problem description is not clear and logic has problems, we conduct manual correction.
\input{figures/prompt}

\textbf{Test Cases Generation. }
Test cases are crucial to validate the correctness and robustness of the code generated by MLLMs. For HumanEval-V, we utilize the test cases provided in HumanEval. For Algorithm, we utilize the test cases from LeetCode dataset\cite{guo2024deepseekcoderlargelanguagemodel}. For MATH subset which do not provide private test cases, we generate a comprehensive set of test cases that cover:
\begin{itemize}
    \item Typical Cases: Standard input scenarios that test the general functionality of the code.
    \item Edge Cases: Boundary conditions that might result in unusual outputs, such as extreme values or zero inputs.
    \item Large Number Cases: These cases involve inputs with significantly large values or large datasets to test the efficiency and performance scalability of the generated code. 
\end{itemize}

For each category, we generate three test case inputs separately. Subsequently, we use the correct code to execute based on the input and get the correct output. We keep these correct input-output pairs as test cases.

\subsection{Data Statistics}\label{dataset:sta}
In this subsection, we show the statistics of \benchmark.

As shown in Table \ref{tab: CodeVision stats}, the \benchmark benchmark consists of three subsets: HumanEval-V, Algorithm, and MATH, covering basic programming, algorithmic, and mathematical problem-solving domains, and difficulty levels (Easy, Medium, and Hard). HumanEval-V contains 164 problems with an average of 8.08 test cases and moderate flowchart complexity (10.90 nodes, 11.07 edges). The Algorithm subset comprises 149 problems across three difficulty levels with 100 test cases each, showing increasing flowchart complexity from Easy (12.62 nodes, 13.29 edges) to Hard (18.86 nodes, 19.14 edges) problems. Similarly, the MATH subset includes 125 problems with an average of 8.92 test cases, demonstrating comparable complexity progression from Easy (10.67 nodes, 10.49 edges) to Hard (19.20 nodes, 19.08 edges) problems, indicating that problem difficulty correlates with flowchart complexity across both Algorithm and MATH subsets.

%% file: tables/stats.tex
\begin{table*}[]
\centering
\resizebox{\textwidth}{!}{
\begin{tabular}{l|c|S[table-format=3.2]S[table-format=3.2]S[table-format=3.2]S[table-format=3.2]|S[table-format=3.2]S[table-format=3.2]S[table-format=3.2]S[table-format=3.2]}
\hline
\multirow{2}{*}{\textbf{}} & \multirow{2}{*}{\textbf{HumanEval-V}} & \multicolumn{4}{c|}{\textbf{Algorithm}}                               & \multicolumn{4}{c}{\textbf{MATH}}                                     \\
                           &                                       & \textbf{Easy} & \textbf{Medium} & \textbf{Hard} & \textbf{Total/Avg.} & \textbf{Easy} & \textbf{Medium} & \textbf{Hard} & \textbf{Total/Avg.} \\ \hline
Problem                    & 164.00                                & 45.00         & 75.00           & 29.00         & 149.00                & 45.00         & 40.00           & 40.00         & 125.00                \\
Avg. Test Cases            & \hspace{1em}8.08                                  & 100.00        & 100.00          & 100.00        & 100.00              & 9.04          & 8.95            & 8.78          & 8.92                \\
Avg. Flowchart Nodes       & \hspace{0.5em}10.90                                 & 12.62         & 15.76           & 18.86         & 15.75               & 10.67         & 12.38           & 19.20         & 14.08               \\
Avg. Flowchart Edges       & \hspace{0.5em}11.07                                 & 13.29         & 15.87           & 19.14         & 16.10               & 10.49         & 12.00           & 19.08         & 13.85               \\ \hline
\end{tabular}
}
\caption{\benchmark Statistics. We calculate the average number of test cases, the average number of flowchart edges, and the average number of flowchart nodes. As the difficulty of the problem increases, the average number of edges and nodes increases.}
\label{tab: CodeVision stats}
\end{table*}

%% file: figures/prompt.tex
\begin{figure}[t] \centering
    \includegraphics[width=0.48\textwidth]{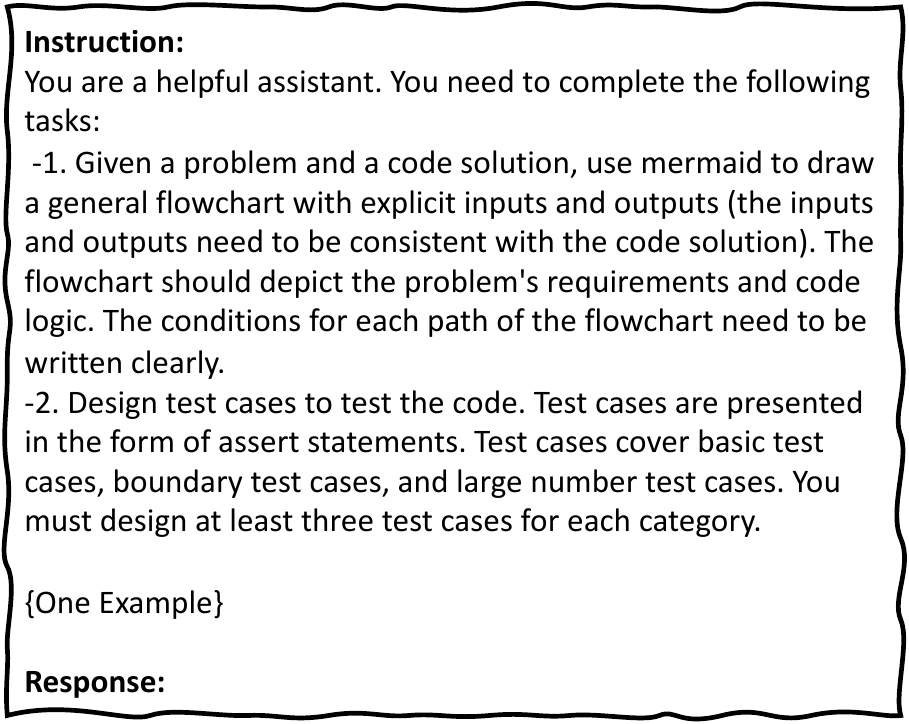}
    \caption{The Prompt Template Used for Flowchart Construction and Test Cases Generation. We provide an example of keeping the output of the model in the expected format. The example is in the Appendix \ref{sec:prompt}.} \label{fig:prompt}
\end{figure}

%% file: sec/4_experimental_methodology.tex
\section{Experimental Methodology}
In this section, we describe the evaluated models, evaluation metrics, and implementation details of our experiments.

\input{tables/overall}
\textbf{Models. }
The models we use can be distinguished as proprietary and open-source models.

For proprietary models, we include GPT-4o~\cite{gpt4o}, the Claude family (Claude 3.5 Sonnet~\cite{Claude3.5_Sonnet}, Claude 3 Sonnet~\cite{Claude3}, Claude 3 Haiku~\cite{Claude3}) and the Gemini family (Gemini 1.5 Pro~\cite{geminiteam2024gemini15unlockingmultimodal}, Gemini 1.5 Flash~\cite{geminiteam2024gemini15unlockingmultimodal}).

For open-source models, we include the Llama-3.2-Vision family (Llama-3.2-11B-Vision-Instruct~\cite{Llama3.2}, Llama-3.2-90B-Vision-Instruct~\cite{Llama3.2}), the Phi-3 family (Phi-3.5-vision-instruct~\cite{abdin2024phi3technicalreporthighly}, Phi-3-vision-128k-instruct~\cite{abdin2024phi3technicalreporthighly}), MiniCPM-V 2.6~\cite{yao2024minicpmvgpt4vlevelmllm} , and Qwen-VL-Plus~\cite{bai2023qwenvlversatilevisionlanguagemodel}.

\textbf{Evaluation Metrics. }
We follow previous work~\cite{chen2021evaluatinglargelanguagemodels,li2024mmcode,wang2024intervenor,yang2024enhancing,luo2023wizardcoder} and we use Pass@$k$ ~\cite{chen2021evaluatinglargelanguagemodels} to evaluate the effectiveness of different MLLMs. Pass@$k$ represents the probability that at least one correct solution appears among the top $k$ generated solutions for each problem:
\begin{equation}
    \text{Pass@}k:=\underset{\text{Problems}}{\operatorname*{\mathbb{E}}}\left[1-\frac{\binom{n-c}k}{\binom nk}\right]
\end{equation}
where $n$ denotes the total number of generated solutions, $c$ is the number of correct solutions, and $k$ is the number of top-ranked solutions being evaluated. In this work, we set $k=1$.

\textbf{Implementation Details. } 
For all MLLMs, we set the generation temperature to 0.2, the nucleus sampling parameter $top\_p$ to 0.95, and the maximum generation length to 1024 tokens, the same as the other code generation work~\cite{hui2024qwen25codertechnicalreport,zheng2024opencodeinterpreter,guo2024deepseek,zhu2024deepseek}. For the proprietary models, we use the API endpoints provided by the respective vendors, and for the open-source models, we use the transformers\footnote{\url{https://huggingface.co/docs/transformers/index}} framework for inference. We use the 0-shot setting in our experiments. The proprietary models we used are gpt-4o-2024-08-06, claude-3-5-sonnet-20240620, claude-3-sonnet-20240229, claude-3-haiku-20240307, Gemini 1.5 Pro (May 2024), and Gemini 1.5 Flash (May 2024).

%% file: tables/overall.tex
\begin{table*}[t]
\centering
\resizebox{\textwidth}{!}{
\begin{tabular}{lcccccccccc}
\hline
\multicolumn{1}{l|}{\multirow{2}{*}{\textbf{Model}}}     & \multicolumn{1}{c|}{\multirow{2}{*}{\textbf{HumanEval-V}}} & \multicolumn{4}{c|}{\textbf{Algorithm}}                      & \multicolumn{4}{c|}{\textbf{MATH}}                            & \multirow{2}{*}{\textbf{Avg.}} \\
\multicolumn{1}{l|}{}                           & \multicolumn{1}{c|}{}                             & \textbf{Easy} & \textbf{Medium} & \textbf{Hard} & \multicolumn{1}{c|}{\textbf{Overall}} & \textbf{Easy}  & \textbf{Medium} & \textbf{Hard} & \multicolumn{1}{c|}{\textbf{Overall}} &                       \\ \hline
\multicolumn{11}{c}{Proprietary Models}                                                                                                                                                                                                  \\ \hline
\multicolumn{1}{l|}{GPT-4o\cite{gpt4o}}                     & \multicolumn{1}{c|}{\textbf{93.9}}                         & \textbf{91.1} & \textbf{89.3}   & \textbf{79.3} & \multicolumn{1}{c|}{\textbf{87.9}}    & \textbf{100.0} & \textbf{97.5}   & \textbf{77.5} & \multicolumn{1}{c|}{\textbf{92.0}}    & \textbf{91.3}                  \\
\multicolumn{1}{l|}{Claude 3.5 Sonnet\cite{Claude3.5_Sonnet}}          & \multicolumn{1}{c|}{82.3}                         & 84.4 & 77.3   & 48.3 & \multicolumn{1}{c|}{73.8}    & 97.8  & 95.0   & 52.5 & \multicolumn{1}{c|}{82.4}    & 79.5                  \\
\multicolumn{1}{l|}{Claude 3 Sonnet\cite{Claude3}}            & \multicolumn{1}{c|}{53.0}                         & 31.1 & 14.7   & 3.4  & \multicolumn{1}{c|}{17.4}    & 88.9  & 72.5   & 17.5 & \multicolumn{1}{c|}{60.8}    & 43.7                  \\
\multicolumn{1}{l|}{Claude 3 Haiku\cite{Claude3}}             & \multicolumn{1}{c|}{48.8}                         & 22.2 & 4.0    & 0.0  & \multicolumn{1}{c|}{8.7}     & 73.3  & 62.5   & 12.5 & \multicolumn{1}{c|}{50.4}    & 37.0                  \\
\multicolumn{1}{l|}{Gemini 1.5 Pro\cite{geminiteam2024gemini15unlockingmultimodal}}             & \multicolumn{1}{c|}{89.6}                         & 88.9 & 72.0   & 44.8 & \multicolumn{1}{c|}{71.8}    & 95.6  & 92.5   & 57.5 & \multicolumn{1}{c|}{82.4}    & 81.3                  \\
\multicolumn{1}{l|}{Gemini 1.5 Flash\cite{geminiteam2024gemini15unlockingmultimodal}}           & \multicolumn{1}{c|}{72.0}                         & 40.0 & 32.0   & 10.3 & \multicolumn{1}{c|}{30.2}    & 86.7  & 72.5   & 20.0 & \multicolumn{1}{c|}{60.8}    & 57.1                  \\ \hline
\multicolumn{11}{c}{Open-source Models}                                                                                                                                                                                                  \\ \hline
\multicolumn{1}{l|}{Llama-3.2-11B-Vision-Ins\cite{Llama3.2}}   & \multicolumn{1}{c|}{29.3}                         & 8.9  & 1.3    & 0.0  & \multicolumn{1}{c|}{3.4}     & 62.2  & 27.5   & 2.5  & \multicolumn{1}{c|}{32.0}    & 23.8                  \\
\multicolumn{1}{l|}{Llama-3.2-90B-Vision-Ins\cite{Llama3.2}}   & \multicolumn{1}{c|}{\textbf{40.9}}                & 17.8 & 8.0    & 0.0  & \multicolumn{1}{c|}{9.4}     & \textbf{80.0}  & \textbf{75.0}   & \textbf{15.0} & \multicolumn{1}{c|}{\textbf{57.6}}    & \textbf{36.9}                 \\
\multicolumn{1}{l|}{Phi-3-vision-128k-instruct\cite{abdin2024phi3technicalreporthighly}} & \multicolumn{1}{c|}{29.3}                         & 15.6 & 4.0    & 0.0  & \multicolumn{1}{c|}{6.7}     & 35.6  & 7.5    & 2.5  & \multicolumn{1}{c|}{16.0}    & 19.6                  \\
\multicolumn{1}{l|}{Phi-3.5-vision-instruct\cite{abdin2024phi3technicalreporthighly}}    & \multicolumn{1}{c|}{28.0}                         & 8.9  & 0.0    & 0.0  & \multicolumn{1}{c|}{2.7}     & 33.3  & 7.5    & 0.0  & \multicolumn{1}{c|}{14.4}    & 17.7                  \\
\multicolumn{1}{l|}{MiniCPM-V 2.6\cite{yao2024minicpm}}              & \multicolumn{1}{c|}{40.2}                         & \textbf{22.2} & \textbf{9.3}    & 0.0  & \multicolumn{1}{c|}{\textbf{11.4}}    & 46.7  & 15.0   & 2.5  & \multicolumn{1}{c|}{22.4}    & 27.8                  \\
\multicolumn{1}{l|}{Qwen-VL-Plus\cite{bai2023qwenvlversatilevisionlanguagemodel}}               & \multicolumn{1}{c|}{17.1}                         & 11.1 & 0.0    & 0.0  & \multicolumn{1}{c|}{3.4}     & 13.3  & 2.5    & 0.0  & \multicolumn{1}{c|}{5.6}     & 11.7                  \\ \hline
\end{tabular}
}
\caption{Overall Performance of MLLMs on \benchmark. Proprietary Models and open-source models show significant performance differences.}
\label{tab: overall}
\end{table*}

%% file: sec/5_evaluation_results.tex
\section{Evaluation Results}

In this section, we first benchmark Multimodal LLMs on \benchmark. Subsequently, we compare the advantages of \benchmark over MMCode and how it can evaluate the reasoning capabilities of MLLMs from different perspectives compared to MathVista. Additionally, we show that MLLMs struggle to understand visually complex logic for code generation and analyze the reasons for the poor performance of open-source MLLMs on \benchmark. Finally, case studies are presented.

\subsection{Overall Performance}
Table \ref{tab: overall} shows the performance of the MLLMs on \benchmark dataset.

Overall, the proprietary models significantly outperform the open-source models. GPT-4o leads the proprietary models with an average score of 91.3, excelling in both the Algorithm and MATH categories. Following GPT-4o, Claude 3.5 Sonnet and Gemini 1.5 Pro also show robust results. In contrast, the open-source models lag behind in terms of overall performance, with Llama-3.2-90B-Vision-Ins scoring the highest among them at 36.9 on average. 
The gap between proprietary and open-source models is especially evident in the Algorithm hard and MATH hard categories, where proprietary models, particularly GPT-4o and Claude 3.5 Sonnet, maintain relatively high scores compared to their open-source counterparts. This disparity suggests that while open-source models may handle simpler tasks effectively, they struggle with more complex problem-solving and reasoning challenges.

Additionally, we observe that the MLLMs score the lowest on the Algorithm subset on average, and all open-source models fail in all questions of Algorithm Hard. This suggests that the problems in the Algorithm subset are more challenging for the MLLMs, which demonstrates that open-source MLLMs still have a significant gap compared to proprietary models in more complex reasoning tasks.

\subsection{Comparision with Other Reasoning Benchmarks MMCode and MathVisita}
\input{tables/compare_mmcode}
\input{figures/compare_mathvista}
In this subsection, we show the advantages that \benchmark has over the other multimodal reasoning benchmarks MMCode and MathVisita.

\textbf{Compare with MMCode. }MMCode is a multimodal reasoning benchmark that evaluates algorithmic problem-solving skills in visually rich contexts. However, a major issue with MMCode is that most of its problems can be solved without using images, making it difficult to determine whether the model utilizes visual information in its responses, which limits its effectiveness in evaluating the reasoning abilities of MLLMs. \benchmark addresses this issue effectively by using flowcharts as a primary input type, making it challenging for models to provide correct answers without the images.

We explore model performance in both the ``Text Only'' and ``Text + Image'' modes to validate \benchmark’s reliance on visual information. As shown in Table ~\ref{tab: compare_mmcode}, on \benchmark, the scores of GPT-4o and Gemini 1.5 Pro in the ``Text Only'' mode are significantly lower than in the ``Text + Image'' mode, indicating a notable performance difference. This difference suggests that \benchmark’s problem design effectively encourages the use of multimodal information, making image input critical for problem-solving. In contrast, in MMCode, the score difference between the two modes is relatively small, indicating that most MMCode problems can be answered solely based on text information, without the need for images. This design limits MMCode’s ability to fully reflect the multimodal reasoning capabilities of MLLMs. By introducing complex image information, such as flowcharts, \benchmark requires models to rely on visual understanding to answer correctly, thus providing a more accurate assessment of the model's multimodal reasoning capabilities.

\textbf{Compare with MathVista. }MathVista is a multimodal mathematics reasoning benchmark. We select some proprietary models and open-source models with similar performance on MathVista to test their performance on \benchmark, to demonstrate the differences between \benchmark and multimodal mathematical reasoning datasets.

As shown in Figure ~\ref{fig:compare_mathvista}, for proprietary models, their performance on \benchmark and MathVista is similar. However, for open-source models, their performance on \benchmark is significantly lower than on MathVista, with a performance gap of around -30\%. This indicates that \benchmark can evaluate the reasoning abilities of MLLMs from different perspectives, such as algorithmic logic understanding, further revealing the shortcomings and limitations of these models in handling complex reasoning tasks.

\input{tables/compare_mermaid}
\input{figures/error_analysis}
\subsection{MLLMs Struggle to Understand Visually Complex Logic for Code Generation}
In this subsection, we compare the performance of the MLLMs when flowchart or mermaid are used as input to demonstrate that MLLMs struggle to understand complex visual logic for code generation.

As shown in Table \ref{tab: mermaid}, when using mermaid as input, most models exhibit performance improvements. This suggests that the mermaid, being a more structured and textual representation, may be easier for MLLMs to interpret compared to flowcharts, which require the understanding of more complex visual elements. In particular, the performance of open-source models on medium and hard problems shows substantial improvements with mermaid inputs, this highlights the challenges MLLMs face in perceiving and processing visually complex logic, which is crucial for accurate code generation.

\subsection{Error Analysis and Case Studies}

Figure \ref{fig:error_analysis} shows the proportion of error types in the code generated by proprietary and open-source models across the three subsets of the \benchmark.

From the pie charts, we can observe that proprietary models have a significantly higher proportion of \texttt{AssertionError} compared to open-source models. In contrast, open-source models have a higher proportion of errors on \texttt{TypeError}, \texttt{IndentationError}, \texttt{SyntaxError}, \texttt{ValueError}, and \texttt{NameError}, with the highest proportion occurring in \texttt{SyntaxError}. This indicates that proprietary models adhere to basic syntax rules more effectively and are able to comprehend the logic within flowcharts. However, open-source models lack this refinement, their generated code has over a 10\% likelihood of failing to compile, indicating that their coding capabilities are poor, even to the point of failing to generate executable code. In our opinion, the programming ability of MLLMs is an essential skill for the path to AGI, as it represents a crucial intersection of logical reasoning, problem-solving,  and the ability to generate precise, executable instructions.

Finally, we conduct case studies to examine the performance of the models on the \benchmark. As shown in Figure \ref{fig:case}, the problem has two flowcharts and puts higher demands on the capabilities of MLLMs. GPT-4o correctly follows the logic depicted in the flowchart to generate the correct code. Llama-3.2-90B-Vision-Instruct understands the intent of the flowchart and uses a dynamic programming algorithm. MiniCPM-V-2\_6 omits two steps from the subgraph and Claude-3-Haiku-20240307 fails to correctly identify the function variables in both the main and sub-functions and does not properly understand the execution flow of the code in the subgraph. More details are shown in Appendix \ref{sec:cases}

%% file: tables/compare_mmcode.tex
\begin{table}[]
\centering
\resizebox{0.48\textwidth}{!}{
\begin{tabular}{l|cc|cc}
\hline
\multirow{2}{*}{\textbf{Model}} & \multicolumn{2}{c|}{\textbf{Code-Vision}}  & \multicolumn{2}{c}{\textbf{MMCode}}        \\
                                & \textbf{Text Only} & \textbf{Text + Image} & \textbf{Text Only} & \textbf{Text + Image} \\ \hline
GPT-4o                          & 24.8               & 87.9                  & 14.8               & 17.0                  \\
Gemini 1.5 Pro                  & 11.4               & 71.8                  & 5.7                & 5.0                   \\ \hline
\end{tabular}
}
\caption{Comparision with MMCode Benchmark. ``Text Only'' and ``Text + Image'' mean that the model input only contains problem description and the input contains a description of the problem and images respectively.}
\label{tab: compare_mmcode}
\vspace{-10pt}
\end{table}

%% file: figures/compare_mathvista.tex
\begin{figure*}[t] \centering
    \includegraphics[width=1\textwidth]{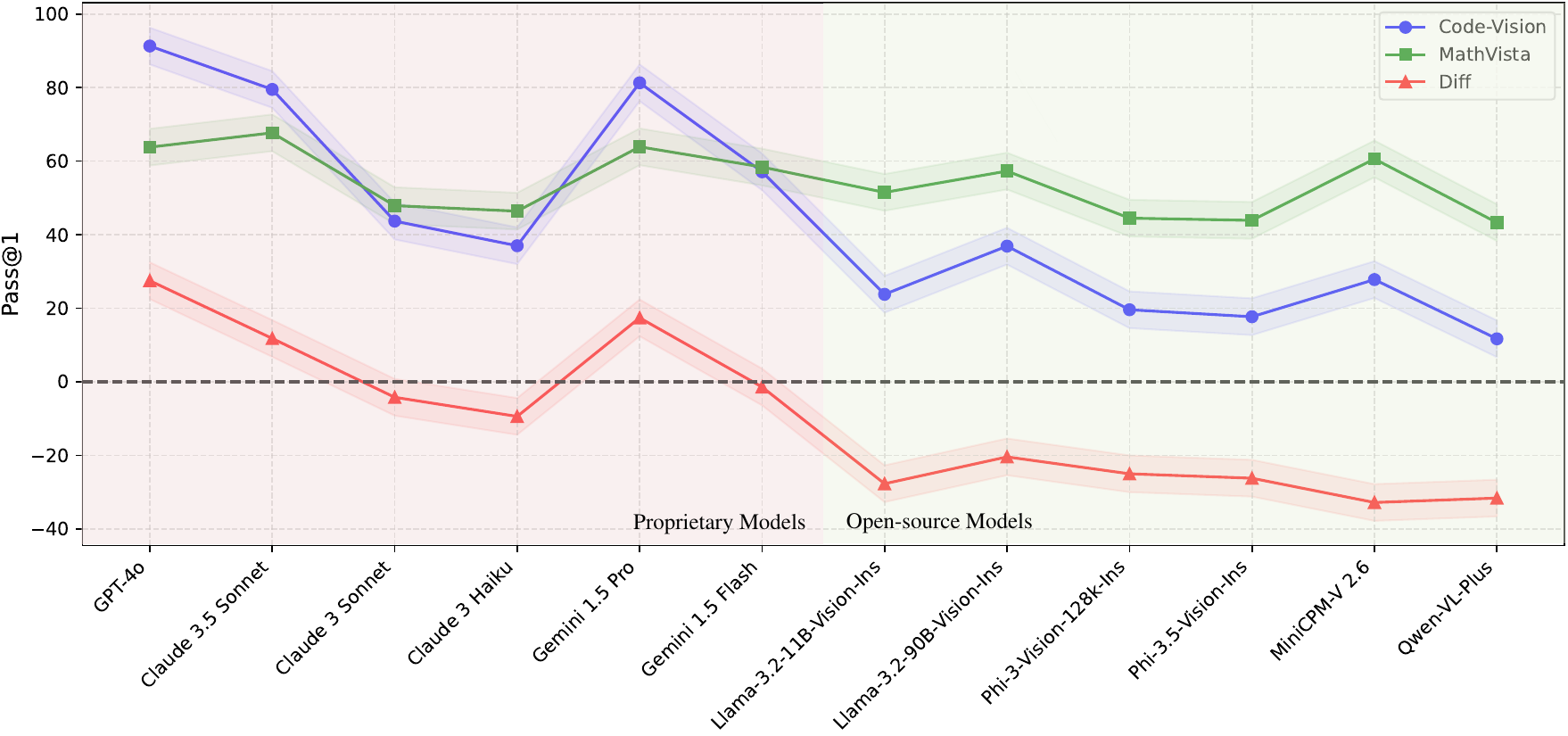}
    \caption{Comparision with MathVista Benchmark. All MLLMs perform similarly on MathVista but have large differences on \benchmark. The diff is the performance of the model on \benchmark minus the performance on MathVista. Detailed results are in appendix \ref{sec:detail_results}} \label{fig:compare_mathvista}
\end{figure*}

%% file: tables/compare_mermaid.tex
\begin{table*}[t]
\centering
\resizebox{\textwidth}{!}{
\begin{tabular}{lcccccc}
\hline
                                 & \multicolumn{2}{c}{\textbf{Easy}}              & \multicolumn{2}{c}{\textbf{Medium}} & \multicolumn{2}{c}{\textbf{Hard}} \\
\multirow{-2}{*}{\textbf{Model}} & Flowchart & Mermaid                            & Flowchart       & Mermaid           & Flowchart      & Mermaid          \\ \hline
\multicolumn{7}{c}{Proprietary Models}                                                                                                                      \\ \hline
GPT-4o                           & 91.1      &  88.9 {\color[HTML]{FE0000}(-2.2)} & 89.3            & 85.3 {\color[HTML]{FE0000}(-4.0)}       & 79.3           & 82.8 {\color[HTML]{009901}(+3.5)}      \\
Claude 3.5 Sonnet                & 84.4      & 82.2 {\color[HTML]{FE0000}(-2.2)}                        & 77.3            & 73.3 {\color[HTML]{FE0000}(-4.0)}       & 48.3           & 44.8 {\color[HTML]{FE0000}(-3.5)}      \\
Claude 3 Sonnet                  & 31.1      & 37.8 {\color[HTML]{009901}(+6.7)} & 14.7            & 20.0 {\color[HTML]{009901}(+5.3)}       & 3.4            & 17.2 {\color[HTML]{009901}(+13.8)}     \\
Claude 3 Haiku                   & 22.2      & 22.2 {\color[HTML]{009901}(+0.0)}                        & 4.0             & 10.7 {\color[HTML]{009901}(+6.7)}       & 0.0            & 6.9 {\color[HTML]{009901}(+6.9)}       \\
Gemini 1.5 Pro                   & 88.9      & 84.4 {\color[HTML]{FE0000}(-4.5)}                        & 72.0            & 73.3 {\color[HTML]{009901}(+1.3)}       & 44.8           & 51.7 {\color[HTML]{009901}(+6.9)}      \\
Gemini 1.5 Flash                 & 40.0      & 44.4 {\color[HTML]{009901}(+4.4)}                        & 32.0            & 44.0 {\color[HTML]{009901}(+12.0)}      & 10.3           & 20.7 {\color[HTML]{009901}(+10.4)}     \\ \hline
\multicolumn{7}{c}{Open-source Models}                                                                                                                      \\ \hline
Llama-3.2-11B-Vision-Instruct    & 8.9       & 28.9 {\color[HTML]{009901}(+20.0)}                       & 1.3             & 6.7 {\color[HTML]{009901}(+5.4)}        & 0.0            & 6.9 {\color[HTML]{009901}(+6.9)}       \\
Llama-3.2-90B-Vision-Instruct    & 17.8      & 28.9 {\color[HTML]{009901}(+11.1)}                       & 8.0             & 13.3 {\color[HTML]{009901}(+5.3)}       & 0.0            & 13.8 {\color[HTML]{009901}(+13.8)}     \\
Phi-3-vision-128k-instruct       & 15.6      & 31.1 {\color[HTML]{009901}(+15.5)}                       & 4.0             & 10.7 {\color[HTML]{009901}(+6.7)}       & 0.0            & 6.9 {\color[HTML]{009901}(+6.9)}       \\
Phi-3.5-vision-instruct          & 8.9       & 22.2 {\color[HTML]{009901}(+13.3)}                       & 0.0             & 9.3 {\color[HTML]{009901}(+9.3)}        & 0.0            & 3.5 {\color[HTML]{009901}(+3.4)}       \\
MiniCPM-V 2.6                    & 22.2      & 33.3 {\color[HTML]{009901}(+11.1)}                       & 9.3             & 18.7 {\color[HTML]{009901}(+9.4)}       & 0.0            & 6.9 {\color[HTML]{009901}(+6.9)}       \\
Qwen-VL-Plus                     & 11.1      & 31.1 {\color[HTML]{009901}(+20.0)}                       & 0.0             & 9.3 {\color[HTML]{009901}(+9.3)}        & 0.0            & 10.3 {\color[HTML]{009901}(+10.3)}     \\ \hline
\end{tabular}
}
\caption{Comparison of MLLMs When Flowchart or Mermaid Are Used as Input. We calculate the performance change ($Flowchart-Mermaid$), with {\color[HTML]{FE0000}red} representing a drop and {\color[HTML]{009901}green} representing a rise. Open-source models show substantial improvements with mermaid inputs. We use Algorithm subset.}
\label{tab: mermaid}
\end{table*}

%% file: figures/error_analysis.tex
\begin{figure*}[t] \centering
    \includegraphics[width=\textwidth]{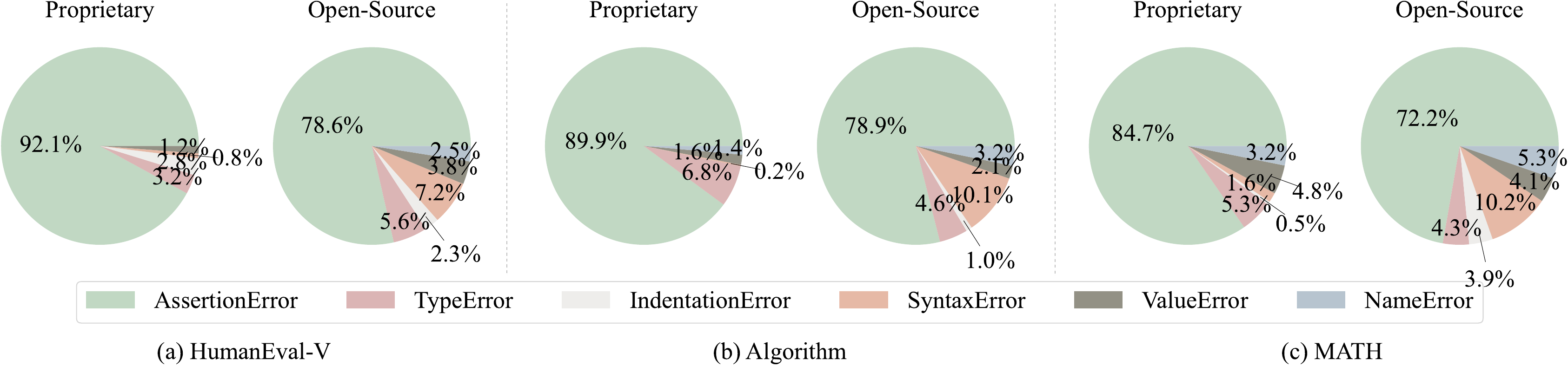}
    \caption{Error Analysis of Proprietary Models and Open-source Models. We count the proportion of each error type in the code generated by proprietary and open-source Models.} \label{fig:error_analysis}
\end{figure*}

%% file: sec/6_conclusion.tex
\section{Conclusion}
We present \benchmark, a novel benchmark for evaluating the logical understanding and code generation capabilities of Multimodal Large Language Models. \benchmark's unique visual-centric design using flowcharts as primary inputs provides a more rigorous test of models' multimodal reasoning abilities compared to existing benchmarks. Our extensive experiments reveal significant performance gaps between proprietary and open-source models. Moreover, we show that the \benchmark can pose unique challenges compared to other multimodal reasoning benchmarks and further demonstrate that open-source models struggle to understand visually complex logic. In the future, we will build more and harder problems that contain more complex logic and flowcharts to challenge proprietary models.

%% file: sec/7_Limitations.tex
The dataset size of \benchmark is relatively limited, especially in the Hard category of the Algorithm and MATH subsets, which may limit the evaluation of the model on extreme cases or rare problems. The evaluation focuses on the correctness of the code and does not adequately evaluate the quality of the generated code, such as the readability, efficiency, and style of the code.

%% file: sec/8_Ethics.tex
Throughout the entire research process and in presenting the findings in this paper, we have maintained strict adherence to ethical standards. Our dataset is established based on commonly used datasets and authoritative platforms, and the relevant data and code have undergone stringent ethical review.

%% file: sec/X_suppl.tex
\section{Appendix}


\input{tables/compare_mathvista}
\subsection{Detail Results}
\label{sec:detail_results}
Table \ref{tab: mathvista} shows detailed experimental results for MLLMs on \benchmark and MathVista.

\subsection{Prompts}
\label{sec:prompt}
The prompts used are listed in Table \ref{tab:prompt_construction1}-\ref{tab:evaluation}.
\begin{table*}[htbp]
\centering
\captionsetup{justification=centering}
\begin{tabular}{p{\textwidth}}
\toprule
\underline{\textbf{\textsc{Prompt for Flowchart Construction and Test Cases Generation.}}} \\
\begin{minipage}{\textwidth}
\vspace{2mm}
\begin{minted}[fontsize=\small,breaklines=true]{markdown}

You are a helpful assistant. You need to complete the following tasks:
1.Given a problem and a code solution, use mermaid to draw a general flowchart with explicit inputs and outputs (the inputs and outputs need to be consistent with the code solution). The flowchart should depict the problem's requirements and code logic. The conditions for each path of the flowchart need to be written clearly. Note the syntax correctness of mermaid, such as the correct use of (), etc. For example, if there are special symbols in mermaid, you need to use "" to include, for example B[Input: x (integer)] is wrong, you need to use B["Input: x (integer)"]. The flowchart is enclosed with ```mermaid and ```.
    -The logic of the flowchart needs to be exactly the same as the code
    -If there is function nesting, display the nested functions as a subgraph
2.Design test cases to test the code. Test cases are presented in the form of assert statements. Test cases cover basic test cases, boundary test cases, and large number test cases. You must design at least three test cases for each category. The test cases are contained with ```python and ```.

<problem>
Palindrome Number
Given an integer x, return true if x is a palindrome, and false otherwise.

Example 1:

Input: x = 121
Output: true
Explanation: 121 reads as 121 from left to right and from right to left.
Example 2:

Input: x = -121
Output: false
Explanation: From left to right, it reads -121. From right to left, it becomes 121-. Therefore it is not a palindrome.
Example 3:

Input: x = 10
Output: false
Explanation: Reads 01 from right to left. Therefore it is not a palindrome.

Constraints:

-231 <= x <= 231 - 1
</problem>

<solution>
def isPalindrome(self, x: int) -> bool:
    if x < 0:
        return False

    reversed_num = 0
    temp = x

    while temp != 0:
        digit = temp % 10
        reversed_num = reversed_num * 10 + digit
        temp //= 10

    return reversed_num == x
</solution>

\end{minted}
\end{minipage} \\
\bottomrule
\end{tabular}

\caption{Prompt for Flowchart Construction and Test Cases Generation.}
\label{tab:prompt_construction1}
\end{table*}

\begin{table*}[htbp]
\centering
\captionsetup{justification=centering}
\begin{tabular}{p{\textwidth}}
\toprule
\underline{\textbf{\textsc{Prompt for Flowchart Construction and Test Cases Generation.}}} \\
\begin{minipage}{\textwidth}
\vspace{2mm}
\begin{minted}[fontsize=\small,breaklines=true]{markdown}
```mermaid
flowchart TD
    A[Start] --> B["Input: x (integer)"]
    B --> C{Is x < 0?}
    C -- Yes --> D["Output: False (negative numbers are not palindromes)"]
    C -- No --> E[Initialize reversed_num = 0, temp = x]
    E --> F{Is temp != 0?}
    F -- Yes --> G[Extract last digit: digit = temp % 10]
    G --> H[Update reversed_num: reversed_num = reversed_num * 10 + digit]
    H --> I[Update temp: temp = temp // 10]
    I --> F
    F -- No --> J{Is reversed_num == x?}
    J -- Yes --> K["Output: True (x is a palindrome)"]
    J -- No --> L["Output: False (x is not a palindrome)"]
```

```python
# Basic test cases
assert isPalindrome(121) == True  # palindrome
assert isPalindrome(-121) == False  # negative number, not a palindrome
assert isPalindrome(10) == False  # not a palindrome

# Boundary test cases
assert isPalindrome(0) == True  # smallest non-negative integer, palindrome
assert isPalindrome(9) == True  # single-digit number, palindrome
assert isPalindrome(-1) == False  # smallest negative number, not a palindrome

# Large number test cases
assert isPalindrome(123454321) == True  # large palindrome
assert isPalindrome(1000000001) == True  # large palindrome with zeros in the middle
assert isPalindrome(123456789) == False  # large non-palindrome
```

<problem>
%%%problem%%%
</problem>

<solution>
%%%solution%%%
</solution>

\end{minted}
\end{minipage} \\
\bottomrule
\end{tabular}

\caption{Prompt for Flowchart Construction and Test Cases Generation.}
\label{tab:prompt_construction2}
\end{table*}

\begin{table*}[htbp]
\centering
\captionsetup{justification=centering}
\begin{tabular}{p{\textwidth}}
\toprule
\underline{\textbf{\textsc{Prompt for Evaluation.}}} \\
\begin{minipage}{\textwidth}
\vspace{2mm}
\begin{minted}[fontsize=\small,breaklines=true]{markdown}
Generate code according to flowchart.
Note: If you're using a specific python package, you'll need to import it yourself. You don't need to use functions like input() to get input, just complete the python function.
Starter Code:
```python
%%%starter_code%%%
```
Present the code between ```python and ```.

\end{minted}
\end{minipage} \\
\bottomrule
\end{tabular}

\caption{Prompt for Evaluation.}
\label{tab:evaluation}
\end{table*}

\subsection{Cases}
\label{sec:cases}
Finally, as shown in Figure \ref{fig:case}, we select the problem from the Math subset and present the responses from four MLLMs. This problem has two flowcharts and puts higher demands on the capabilities of MLLMs.

Overall, GPT-4o and Llama-3.2-90B-Vision-Instruct provide correct answers, while MiniCPM-V-2\_6 and Claude-3-Haiku-20240307 gave incorrect ones. Upon careful analysis of the generated code, we can observe that GPT-4o correctly follows the logic depicted in the flowchart to generate the correct code. However, Llama-3.2-90B-Vision-Instruct did not follow the recursive logic as shown in the flowchart, instead using the dynamic programming algorithm. This is an interesting phenomenon, suggesting that Llama-3.2-90B-Vision-Instruct, after processing the flowchart, comprehends the problem that the algorithm in the chart is solving and provides a dynamic programming solution. This demonstrates that the model not only interprets the flowchart's surface meaning but also, to some extent, ``understands" the logical intent behind it.

As clearly shown in the bottom of Figure \ref{fig:case}, MiniCPM-V-2\_6 fails to generate fully correct code because the model omits two steps from the subgraph, although the other steps are correct, ultimately leading to an \texttt{AssertionError}. On the other hand, Claude-3-Haiku-20240307 fails to correctly identify the function variables in both the main and sub-functions and does not properly understand the execution flow of the code in the subgraph, ultimately leading to a \texttt{RecursionError}. This indicates that both models have poor image understanding and reasoning capabilities.
\input{figures/case}

%% file: tables/compare_mathvista.tex
\begin{table}[t]
\centering
\resizebox{0.48\textwidth}{!}{
\begin{tabular}{lcc}
\hline
\textbf{Model}           & \textbf{Code-Vision} & \textbf{MathVista} \\ \hline
GPT-4o                   & 91.3                 & 63.8               \\
Claude 3.5 Sonnet        & 79.5                 & 67.7               \\
Claude 3 Sonnet          & 43.7                 & 47.9               \\
Claude 3 Haiku           & 37.0                 & 46.4               \\
Gemini 1.5 Pro           & 81.3                 & 63.9               \\
Gemini 1.5 Flash         & 57.1                 & 58.4               \\
Llama-3.2-11B-Vision-Ins & 23.8                 & 51.5               \\
Llama-3.2-90B-Vision-Ins & 36.9                 & 57.3               \\
Phi-3-Vision-128k-Ins    & 19.6                 & 44.5               \\
Phi-3.5-Vision-Ins       & 17.7                 & 43.9               \\
MiniCPM-V 2.6            & 27.8                 & 60.6               \\
Qwen-VL-Plus             & 11.7                 & 43.3               \\ \hline
\end{tabular}
}
\caption{Detailed results of MLLMs on \benchmark and MathVista.}
\label{tab: mathvista}
\end{table}

%% file: figures/case.tex
\begin{figure*}[t] \centering
    \includegraphics[width=0.8\textwidth]{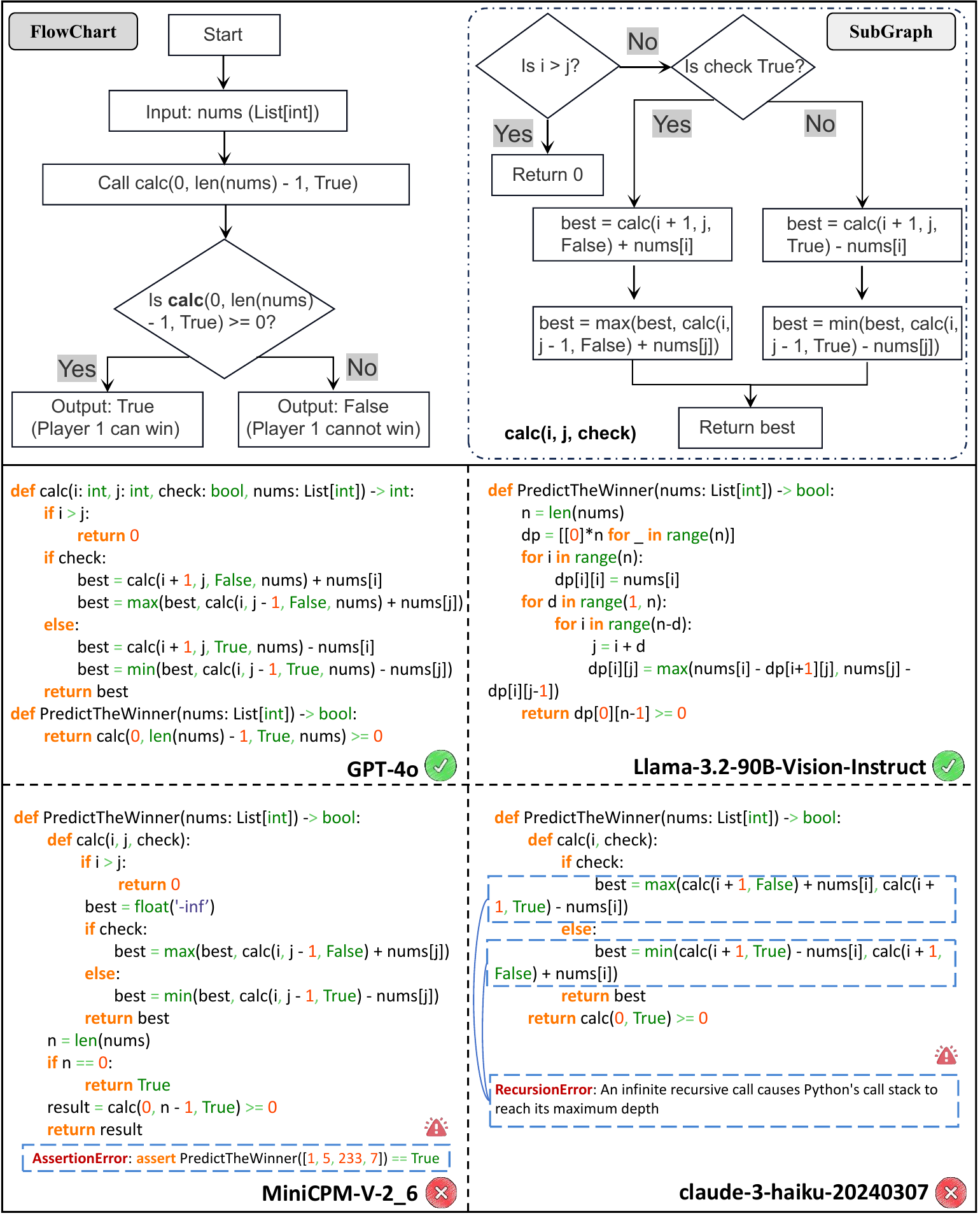}
    \caption{Case Studies. We provide a flowchart of the problem (with 1 subgraph) from the Math subset and present the responses from four MLLMs.} \label{fig:case}
\end{figure*}

%% file: acl_latex.bbl
\begin{thebibliography}{42}
\providecommand{\natexlab}[1]{#1}

\bibitem[{Abdin et~al.(2024{\natexlab{a}})Abdin, Aneja, Awadalla, Awadallah, Awan, Bach, Bahree, Bakhtiari, Bao, Behl et~al.}]{abdin2024phi}
Marah Abdin, Jyoti Aneja, Hany Awadalla, Ahmed Awadallah, Ammar~Ahmad Awan, Nguyen Bach, Amit Bahree, Arash Bakhtiari, Jianmin Bao, Harkirat Behl, et~al. 2024{\natexlab{a}}.
\newblock Phi-3 technical report: A highly capable language model locally on your phone.
\newblock \emph{arXiv preprint arXiv:2404.14219}.

\bibitem[{Abdin et~al.(2024{\natexlab{b}})Abdin, Aneja, Awadalla et~al.}]{abdin2024phi3technicalreporthighly}
Marah Abdin, Jyoti Aneja, Hany Awadalla, et~al. 2024{\natexlab{b}}.
\newblock \href {https://arxiv.org/abs/2404.14219} {Phi-3 technical report: A highly capable language model locally on your phone}.
\newblock \emph{Preprint}, arXiv:2404.14219.

\bibitem[{AI(2024)}]{Llama3.2}
Meta AI. 2024.
\newblock \href {https://ai.meta.com/blog/llama-3-2-connect-2024-vision-edge-mobile-devices} {Llama 3.2 vision family}.

\bibitem[{Anthropic(2024{\natexlab{a}})}]{Claude3}
Anthropic. 2024{\natexlab{a}}.
\newblock \href {https://www.anthropic.com/news/claude-3-family} {Claude 3 opus}.

\bibitem[{Anthropic(2024{\natexlab{b}})}]{Claude3.5_Sonnet}
Anthropic. 2024{\natexlab{b}}.
\newblock \href {https://www.anthropic.com/news/claude-3-5-sonnet} {Claude 3.5 sonnet}.

\bibitem[{Austin et~al.(2021)Austin, Odena, Nye, Bosma, Michalewski, Dohan, Jiang, Cai, Terry, Le et~al.}]{austin2021program}
Jacob Austin, Augustus Odena, Maxwell Nye, Maarten Bosma, Henryk Michalewski, David Dohan, Ellen Jiang, Carrie Cai, Michael Terry, Quoc Le, et~al. 2021.
\newblock Program synthesis with large language models.
\newblock \emph{arXiv preprint arXiv:2108.07732}.

\bibitem[{Bai et~al.(2023)Bai, Bai, Yang, Wang, Tan, Wang, Lin, Zhou, and Zhou}]{bai2023qwenvlversatilevisionlanguagemodel}
Jinze Bai, Shuai Bai, Shusheng Yang, Shijie Wang, Sinan Tan, Peng Wang, Junyang Lin, Chang Zhou, and Jingren Zhou. 2023.
\newblock \href {https://arxiv.org/abs/2308.12966} {Qwen-vl: A versatile vision-language model for understanding, localization, text reading, and beyond}.
\newblock \emph{Preprint}, arXiv:2308.12966.

\bibitem[{Chen et~al.(2021)Chen, Tworek, Jun, Yuan, de~Oliveira~Pinto, Kaplan, Edwards, Burda, Joseph, Brockman, Ray, Puri, Krueger, Petrov, Khlaaf, Sastry, Mishkin, Chan, Gray, Ryder, Pavlov, Power, Kaiser, Bavarian, Winter, Tillet, Such, Cummings, Plappert, Chantzis, Barnes, Herbert-Voss, Guss, Nichol, Paino, Tezak, Tang, Babuschkin, Balaji, Jain, Saunders, Hesse, Carr, Leike, Achiam, Misra, Morikawa, Radford, Knight, Brundage, Murati, Mayer, Welinder, McGrew, Amodei, McCandlish, Sutskever, and Zaremba}]{chen2021evaluatinglargelanguagemodels}
Mark Chen, Jerry Tworek, Heewoo Jun, Qiming Yuan, Henrique~Ponde de~Oliveira~Pinto, Jared Kaplan, Harri Edwards, Yuri Burda, Nicholas Joseph, Greg Brockman, Alex Ray, Raul Puri, Gretchen Krueger, Michael Petrov, Heidy Khlaaf, Girish Sastry, Pamela Mishkin, Brooke Chan, Scott Gray, Nick Ryder, Mikhail Pavlov, Alethea Power, Lukasz Kaiser, Mohammad Bavarian, Clemens Winter, Philippe Tillet, Felipe~Petroski Such, Dave Cummings, Matthias Plappert, Fotios Chantzis, Elizabeth Barnes, Ariel Herbert-Voss, William~Hebgen Guss, Alex Nichol, Alex Paino, Nikolas Tezak, Jie Tang, Igor Babuschkin, Suchir Balaji, Shantanu Jain, William Saunders, Christopher Hesse, Andrew~N. Carr, Jan Leike, Josh Achiam, Vedant Misra, Evan Morikawa, Alec Radford, Matthew Knight, Miles Brundage, Mira Murati, Katie Mayer, Peter Welinder, Bob McGrew, Dario Amodei, Sam McCandlish, Ilya Sutskever, and Wojciech Zaremba. 2021.
\newblock \href {https://arxiv.org/abs/2107.03374} {Evaluating large language models trained on code}.
\newblock \emph{Preprint}, arXiv:2107.03374.

\bibitem[{Dai et~al.(2023)Dai, Li, Li, Tiong, Zhao, Wang, Li, Fung, and Hoi}]{dai2023instructblipgeneralpurposevisionlanguagemodels}
Wenliang Dai, Junnan Li, Dongxu Li, Anthony Meng~Huat Tiong, Junqi Zhao, Weisheng Wang, Boyang Li, Pascale Fung, and Steven Hoi. 2023.
\newblock \href {https://arxiv.org/abs/2305.06500} {Instructblip: Towards general-purpose vision-language models with instruction tuning}.
\newblock \emph{Preprint}, arXiv:2305.06500.

\bibitem[{Goyal et~al.(2017)Goyal, Khot, Summers-Stay, Batra, and Parikh}]{goyal2017making}
Yash Goyal, Tejas Khot, Douglas Summers-Stay, Dhruv Batra, and Devi Parikh. 2017.
\newblock Making the v in vqa matter: Elevating the role of image understanding in visual question answering.
\newblock In \emph{Proceedings of the IEEE conference on computer vision and pattern recognition}, pages 6904--6913.

\bibitem[{Guo et~al.(2024{\natexlab{a}})Guo, Zhu, Yang, Xie, Dong, Zhang, Chen, Bi, Wu, Li, Luo, Xiong, and Liang}]{guo2024deepseekcoderlargelanguagemodel}
Daya Guo, Qihao Zhu, Dejian Yang, Zhenda Xie, Kai Dong, Wentao Zhang, Guanting Chen, Xiao Bi, Y.~Wu, Y.~K. Li, Fuli Luo, Yingfei Xiong, and Wenfeng Liang. 2024{\natexlab{a}}.
\newblock \href {https://arxiv.org/abs/2401.14196} {Deepseek-coder: When the large language model meets programming -- the rise of code intelligence}.
\newblock \emph{Preprint}, arXiv:2401.14196.

\bibitem[{Guo et~al.(2024{\natexlab{b}})Guo, Zhu, Yang, Xie, Dong, Zhang, Chen, Bi, Wu, Li et~al.}]{guo2024deepseek}
Daya Guo, Qihao Zhu, Dejian Yang, Zhenda Xie, Kai Dong, Wentao Zhang, Guanting Chen, Xiao Bi, Yu~Wu, YK~Li, et~al. 2024{\natexlab{b}}.
\newblock Deepseek-coder: When the large language model meets programming--the rise of code intelligence.
\newblock \emph{arXiv preprint arXiv:2401.14196}.

\bibitem[{Gurari et~al.(2018)Gurari, Li, Stangl, Guo, Lin, Grauman, Luo, and Bigham}]{gurari2018vizwiz}
Danna Gurari, Qing Li, Abigale~J Stangl, Anhong Guo, Chi Lin, Kristen Grauman, Jiebo Luo, and Jeffrey~P Bigham. 2018.
\newblock Vizwiz grand challenge: Answering visual questions from blind people.
\newblock In \emph{Proceedings of the IEEE conference on computer vision and pattern recognition}, pages 3608--3617.

\bibitem[{Hendrycks et~al.(2021)Hendrycks, Basart, Kadavath, Mazeika, Arora, Guo, Burns, Puranik, He, Song, and Steinhardt}]{hendrycksapps2021}
Dan Hendrycks, Steven Basart, Saurav Kadavath, Mantas Mazeika, Akul Arora, Ethan Guo, Collin Burns, Samir Puranik, Horace He, Dawn Song, and Jacob Steinhardt. 2021.
\newblock Measuring coding challenge competence with apps.
\newblock \emph{NeurIPS}.

\bibitem[{Hui et~al.(2024)Hui, Yang, Cui, Yang, Liu, Zhang, Liu, Zhang, Yu, Lu, Dang, Fan, Zhang, Yang, Men, Huang, Zheng, Miao, Quan, Feng, Ren, Ren, Zhou, and Lin}]{hui2024qwen25codertechnicalreport}
Binyuan Hui, Jian Yang, Zeyu Cui, Jiaxi Yang, Dayiheng Liu, Lei Zhang, Tianyu Liu, Jiajun Zhang, Bowen Yu, Keming Lu, Kai Dang, Yang Fan, Yichang Zhang, An~Yang, Rui Men, Fei Huang, Bo~Zheng, Yibo Miao, Shanghaoran Quan, Yunlong Feng, Xingzhang Ren, Xuancheng Ren, Jingren Zhou, and Junyang Lin. 2024.
\newblock \href {https://arxiv.org/abs/2409.12186} {Qwen2.5-coder technical report}.
\newblock \emph{Preprint}, arXiv:2409.12186.

\bibitem[{Jain et~al.(2024)Jain, Han, Gu, Li, Yan, Zhang, Wang, Solar-Lezama, Sen, and Stoica}]{jain2024livecodebench}
Naman Jain, King Han, Alex Gu, Wen-Ding Li, Fanjia Yan, Tianjun Zhang, Sida Wang, Armando Solar-Lezama, Koushik Sen, and Ion Stoica. 2024.
\newblock Livecodebench: Holistic and contamination free evaluation of large language models for code.
\newblock \emph{arXiv preprint arXiv:2403.07974}.

\bibitem[{Li et~al.(2024{\natexlab{a}})Li, Zhang, Zhang, Guo, Zhang, Li, Zhang, Liu, and Li}]{li2024llava}
Bo~Li, Kaichen Zhang, Hao Zhang, Dong Guo, Renrui Zhang, Feng Li, Yuanhan Zhang, Ziwei Liu, and Chunyuan Li. 2024{\natexlab{a}}.
\newblock Llava-next: Stronger llms supercharge multimodal capabilities in the wild.

\bibitem[{Li et~al.(2024{\natexlab{b}})Li, Zhang, Guo, Zhang, Li, Zhang, Zhang, Zhang, Li, Liu, and Li}]{li2024llavaonevisioneasyvisualtask}
Bo~Li, Yuanhan Zhang, Dong Guo, Renrui Zhang, Feng Li, Hao Zhang, Kaichen Zhang, Peiyuan Zhang, Yanwei Li, Ziwei Liu, and Chunyuan Li. 2024{\natexlab{b}}.
\newblock \href {https://arxiv.org/abs/2408.03326} {Llava-onevision: Easy visual task transfer}.
\newblock \emph{Preprint}, arXiv:2408.03326.

\bibitem[{Li and Lu(2024)}]{li2024survey}
Jian Li and Weiheng Lu. 2024.
\newblock A survey on benchmarks of multimodal large language models.
\newblock \emph{arXiv preprint arXiv:2408.08632}.

\bibitem[{Li et~al.(2024{\natexlab{c}})Li, Tian, Hu, Luo, and Ma}]{li2024mmcode}
Kaixin Li, Yuchen Tian, Qisheng Hu, Ziyang Luo, and Jing Ma. 2024{\natexlab{c}}.
\newblock Mmcode: Evaluating multi-modal code large language models with visually rich programming problems.
\newblock \emph{arXiv preprint arXiv:2404.09486}.

\bibitem[{Li et~al.(2021)Li, Lei, Gan, and Liu}]{li2021adversarial}
Linjie Li, Jie Lei, Zhe Gan, and Jingjing Liu. 2021.
\newblock Adversarial vqa: A new benchmark for evaluating the robustness of vqa models.
\newblock In \emph{Proceedings of the IEEE/CVF International Conference on Computer Vision}, pages 2042--2051.

\bibitem[{Liu et~al.(2024)Liu, Li, Wu, and Lee}]{liu2024visual}
Haotian Liu, Chunyuan Li, Qingyang Wu, and Yong~Jae Lee. 2024.
\newblock Visual instruction tuning.
\newblock \emph{Advances in neural information processing systems}, 36.

\bibitem[{Lu et~al.(2024{\natexlab{a}})Lu, Liu, Zhang, Wang, Dong, Liu, Sun, Ren, Li, Yang et~al.}]{lu2024deepseek}
Haoyu Lu, Wen Liu, Bo~Zhang, Bingxuan Wang, Kai Dong, Bo~Liu, Jingxiang Sun, Tongzheng Ren, Zhuoshu Li, Hao Yang, et~al. 2024{\natexlab{a}}.
\newblock Deepseek-vl: towards real-world vision-language understanding.
\newblock \emph{arXiv preprint arXiv:2403.05525}.

\bibitem[{Lu et~al.(2024{\natexlab{b}})Lu, Bansal, Xia, Liu, Li, Hajishirzi, Cheng, Chang, Galley, and Gao}]{lu2024mathvista}
Pan Lu, Hritik Bansal, Tony Xia, Jiacheng Liu, Chunyuan Li, Hannaneh Hajishirzi, Hao Cheng, Kai-Wei Chang, Michel Galley, and Jianfeng Gao. 2024{\natexlab{b}}.
\newblock Mathvista: Evaluating mathematical reasoning of foundation models in visual contexts.
\newblock In \emph{International Conference on Learning Representations (ICLR)}.

\bibitem[{Luo et~al.(2023)Luo, Xu, Zhao, Sun, Geng, Hu, Tao, Ma, Lin, and Jiang}]{luo2023wizardcoder}
Ziyang Luo, Can Xu, Pu~Zhao, Qingfeng Sun, Xiubo Geng, Wenxiang Hu, Chongyang Tao, Jing Ma, Qingwei Lin, and Daxin Jiang. 2023.
\newblock Wizardcoder: Empowering code large language models with evol-instruct.
\newblock \emph{arXiv preprint arXiv:2306.08568}.

\bibitem[{Marino et~al.(2019)Marino, Rastegari, Farhadi, and Mottaghi}]{marino2019ok}
Kenneth Marino, Mohammad Rastegari, Ali Farhadi, and Roozbeh Mottaghi. 2019.
\newblock Ok-vqa: A visual question answering benchmark requiring external knowledge.
\newblock In \emph{Proceedings of the IEEE/cvf conference on computer vision and pattern recognition}, pages 3195--3204.

\bibitem[{Mobasher et~al.(2022)Mobasher, Zamaninejad, Hashemi, Nobakhtian, and Eetemadi}]{mobasher2022parsvqa}
Shaghayegh Mobasher, Ghazal Zamaninejad, Maryam Hashemi, Melika Nobakhtian, and Sauleh Eetemadi. 2022.
\newblock Parsvqa-caps: A benchmark for visual question answering and image captioning in persian.
\newblock \emph{people}, 101:404.

\bibitem[{Morris et~al.(2023)Morris, Sohl-Dickstein, Fiedel, Warkentin, Dafoe, Faust, Farabet, and Legg}]{morris2023levels}
Meredith~Ringel Morris, Jascha Sohl-Dickstein, Noah Fiedel, Tris Warkentin, Allan Dafoe, Aleksandra Faust, Clement Farabet, and Shane Legg. 2023.
\newblock Levels of agi: Operationalizing progress on the path to agi.
\newblock \emph{arXiv preprint arXiv:2311.02462}.

\bibitem[{OpenAI(2024)}]{gpt4o}
OpenAI. 2024.
\newblock \href {https://openai.com/index/hello-gpt-4o} {Gpt-4o}.

\bibitem[{Shi et~al.(2024)Shi, Yang, Liu, Shui, Wang, Jing, Xu, Zhu, Li, Zhang et~al.}]{shi2024chartmimic}
Chufan Shi, Cheng Yang, Yaxin Liu, Bo~Shui, Junjie Wang, Mohan Jing, Linran Xu, Xinyu Zhu, Siheng Li, Yuxiang Zhang, et~al. 2024.
\newblock Chartmimic: Evaluating lmm's cross-modal reasoning capability via chart-to-code generation.
\newblock \emph{arXiv preprint arXiv:2406.09961}.

\bibitem[{Si et~al.(2024)Si, Zhang, Yang, Liu, and Yang}]{si2024design2code}
Chenglei Si, Yanzhe Zhang, Zhengyuan Yang, Ruibo Liu, and Diyi Yang. 2024.
\newblock Design2code: How far are we from automating front-end engineering?
\newblock \emph{arXiv preprint arXiv:2403.03163}.

\bibitem[{Singh et~al.(2019)Singh, Natarajan, Shah, Jiang, Chen, Batra, Parikh, and Rohrbach}]{singh2019towards}
Amanpreet Singh, Vivek Natarajan, Meet Shah, Yu~Jiang, Xinlei Chen, Dhruv Batra, Devi Parikh, and Marcus Rohrbach. 2019.
\newblock Towards vqa models that can read.
\newblock In \emph{Proceedings of the IEEE/CVF conference on computer vision and pattern recognition}, pages 8317--8326.

\bibitem[{Team et~al.(2024)}]{geminiteam2024gemini15unlockingmultimodal}
Gemini Team et~al. 2024.
\newblock \href {https://arxiv.org/abs/2403.05530} {Gemini 1.5: Unlocking multimodal understanding across millions of tokens of context}.
\newblock \emph{Preprint}, arXiv:2403.05530.

\bibitem[{Wang et~al.(2024{\natexlab{a}})Wang, Liu, Wang, Cui, Ding, Liu, and Yu}]{wang2024intervenor}
Hanbin Wang, Zhenghao Liu, Shuo Wang, Ganqu Cui, Ning Ding, Zhiyuan Liu, and Ge~Yu. 2024{\natexlab{a}}.
\newblock Intervenor: Prompting the coding ability of large language models with the interactive chain of repair.
\newblock In \emph{Findings of the Association for Computational Linguistics ACL 2024}, pages 2081--2107.

\bibitem[{Wang et~al.(2024{\natexlab{b}})Wang, Pan, Shi, Lu, Zhan, and Li}]{wang2024measuring}
Ke~Wang, Junting Pan, Weikang Shi, Zimu Lu, Mingjie Zhan, and Hongsheng Li. 2024{\natexlab{b}}.
\newblock \href {https://arxiv.org/abs/2402.14804} {Measuring multimodal mathematical reasoning with math-vision dataset}.
\newblock \emph{Preprint}, arXiv:2402.14804.

\bibitem[{Yang et~al.(2024)Yang, Wang, Liu, Li, Yan, Wang, Gu, Yu, Liu, and Yu}]{yang2024enhancing}
Weiqing Yang, Hanbin Wang, Zhenghao Liu, Xinze Li, Yukun Yan, Shuo Wang, Yu~Gu, Minghe Yu, Zhiyuan Liu, and Ge~Yu. 2024.
\newblock Enhancing the code debugging ability of llms via communicative agent based data refinement.
\newblock \emph{arXiv preprint arXiv:2408.05006}.

\bibitem[{Yao et~al.(2024{\natexlab{a}})Yao, Yu, Zhang, Wang, Cui, Zhu, Cai, Li, Zhao, He et~al.}]{yao2024minicpm}
Yuan Yao, Tianyu Yu, Ao~Zhang, Chongyi Wang, Junbo Cui, Hongji Zhu, Tianchi Cai, Haoyu Li, Weilin Zhao, Zhihui He, et~al. 2024{\natexlab{a}}.
\newblock Minicpm-v: A gpt-4v level mllm on your phone.
\newblock \emph{arXiv preprint arXiv:2408.01800}.

\bibitem[{Yao et~al.(2024{\natexlab{b}})Yao, Yu, Zhang, Wang et~al.}]{yao2024minicpmvgpt4vlevelmllm}
Yuan Yao, Tianyu Yu, Ao~Zhang, Chongyi Wang, et~al. 2024{\natexlab{b}}.
\newblock \href {https://arxiv.org/abs/2408.01800} {Minicpm-v: A gpt-4v level mllm on your phone}.
\newblock \emph{Preprint}, arXiv:2408.01800.

\bibitem[{Ye et~al.(2024)Ye, Xu, Xu, Ye, Yan, Zhou, Wang, Hu, Shi, Shi, Li, Xu, Chen, Tian, Qian, Zhang, Huang, and Zhou}]{ye2024mplugowlmodularizationempowerslarge}
Qinghao Ye, Haiyang Xu, Guohai Xu, Jiabo Ye, Ming Yan, Yiyang Zhou, Junyang Wang, Anwen Hu, Pengcheng Shi, Yaya Shi, Chenliang Li, Yuanhong Xu, Hehong Chen, Junfeng Tian, Qi~Qian, Ji~Zhang, Fei Huang, and Jingren Zhou. 2024.
\newblock \href {https://arxiv.org/abs/2304.14178} {mplug-owl: Modularization empowers large language models with multimodality}.
\newblock \emph{Preprint}, arXiv:2304.14178.

\bibitem[{Zheng et~al.(2024)Zheng, Zhang, Shen, Liu, Lin, Fu, Chen, and Yue}]{zheng2024opencodeinterpreter}
Tianyu Zheng, Ge~Zhang, Tianhao Shen, Xueling Liu, Bill~Yuchen Lin, Jie Fu, Wenhu Chen, and Xiang Yue. 2024.
\newblock Opencodeinterpreter: Integrating code generation with execution and refinement.
\newblock \emph{arXiv preprint arXiv:2402.14658}.

\bibitem[{Zhu et~al.(2023)Zhu, Chen, Shen, Li, and Elhoseiny}]{zhu2023minigpt}
Deyao Zhu, Jun Chen, Xiaoqian Shen, Xiang Li, and Mohamed Elhoseiny. 2023.
\newblock Minigpt-4: Enhancing vision-language understanding with advanced large language models.
\newblock \emph{arXiv preprint arXiv:2304.10592}.

\bibitem[{Zhu et~al.(2024)Zhu, Guo, Shao, Yang, Wang, Xu, Wu, Li, Gao, Ma et~al.}]{zhu2024deepseek}
Qihao Zhu, Daya Guo, Zhihong Shao, Dejian Yang, Peiyi Wang, Runxin Xu, Y~Wu, Yukun Li, Huazuo Gao, Shirong Ma, et~al. 2024.
\newblock Deepseek-coder-v2: Breaking the barrier of closed-source models in code intelligence.
\newblock \emph{arXiv preprint arXiv:2406.11931}.

\end{thebibliography}
